\documentclass{article} % For LaTeX2e
\usepackage{iclr2026_conference,times}
\iclrfinalcopy

\usepackage{amsmath,amsfonts,bm}

\def\eqref#1{equation~\ref{#1}}
\def\1{\bm{1}}

\DeclareMathAlphabet{\mathsfit}{\encodingdefault}{\sfdefault}{m}{sl}
\SetMathAlphabet{\mathsfit}{bold}{\encodingdefault}{\sfdefault}{bx}{n}

\usepackage{hyperref}
\usepackage{url}
\usepackage{bm}
\usepackage{graphicx}
\usepackage{caption}
\usepackage{booktabs}
\usepackage{hyperref}
\usepackage{adjustbox}
\usepackage{wrapfig}
\usepackage{pifont}
\usepackage{multirow}
\usepackage[dvipsnames]{xcolor}
\usepackage[table,xcdraw]{xcolor}
\hypersetup{
	colorlinks=true,
	citecolor=teal,
}
\usepackage[ruled,vlined,linesnumbered]{algorithm2e}
\usepackage{tabularx}
\usepackage{enumitem}
\usepackage{array}
\usepackage{listings}
\lstdefinestyle{promptstyle}{
  basicstyle=\ttfamily\small,
  breaklines=true,
  columns=fullflexible,
  frame=single,
  framerule=0.4pt,
  rulecolor=\color{black!20},
  backgroundcolor=\color{black!2},
  tabsize=2,
  upquote=true,
  showstringspaces=false
}
\newcolumntype{Y}{>{\raggedright\arraybackslash}X}
\usepackage{booktabs,array}
\definecolor{LightSkyBlue}{RGB}{100,170,230}
\definecolor{darkgreen}{RGB}{0,100,0}
\newcolumntype{C}[1]{>{\centering\arraybackslash}m{#1}} % centered + wrapped
\renewcommand{\arraystretch}{0.98}
\DontPrintSemicolon
\SetKwInput{KwIn}{Input}
\SetKwInput{KwOut}{Output}
\SetKwFunction{LoadImage}{LoadImage}
\SetKwFunction{EncodePNG}{EncodePNG}
\SetKwFunction{ToLAB}{ToLAB}
\SetKwFunction{DeltaE}{DeltaE\_CIE76}
\SetKwFunction{Clamp}{Clamp}
\SetKwProg{Fn}{Function}{:}{}
\title{\textsc{GUI-Spotlight}: Adaptive Iterative Focus Refinement for Enhanced GUI Visual Grounding}

\author{
    \textbf{Bin Lei}$^{1}${ } \textbf{Nuo Xu}$^{1}${ } \textbf{Ali Payani}$^{2}${ } \textbf{Mingyi Hong}$^1${ } \textbf{Chunhua Liao}$^{3}${ } \textbf{Yu Cao}$^{1}${ } \textbf{Caiwen Ding}$^{1}${ } \\
    $^1$University of Minnesota { }
    $^2$Cisco Research { }
    $^3$Lawrence Livermore National Labs { }\\
    \texttt{lei00126@umn.edu}, 
    \texttt{dingc@umn.edu}
}

\begin{document}

\maketitle
\begin{abstract}
Multimodal large language models (MLLMs) have markedly expanded the competence of graphical user‑interface (GUI) systems, propelling them beyond controlled simulations into complex, real‑world environments across diverse platforms.
However, practical usefulness is still bounded by the reliability of visual grounding, i.e., mapping textual references to exact on-screen elements. This limitation prevents the system from accurately performing pointer‑level actions such as clicking or dragging.
To address it, we introduce \textsc{GUI‑Spotlight}—A model trained for \textit{image-grounded reasoning} that dynamically invokes multiple specialized tools to iteratively narrow its focus to the relevant region of the screen, thereby substantially improving visual grounding accuracy. On the ScreenSpot-Pro benchmark, GUI-Spotlight trained with only $\mathbf{18.5}K$ training samples achieves $\mathbf{52.8}\%$ accuracy, surpassing V2P-7B ($50.6\%$ with $9.6M$ training samples) and GTA-1-7B ($50.1\%$ with $1.56M$ training samples). Code is avaliable at \url{https://github.com/bin123apple/GUI_Spotlight}.
\end{abstract}
\section{Introduction}
Recent rapid advances in multimodal large language models have driven swift progress in GUI agents capable of handling complex tasks on general graphical user interfaces (GUIs)~\citep{xie2024osworld,wu2024copilot}. Nevertheless, current GUI agents still lack robust, fine-grained visual grounding, making it difficult to translate \emph{what} to do into \emph{where} to act on complex, dynamically changing screens~\citep{jang2024videowebarena,xie2025scaling}. As a result, they struggle to reliably perform pixel-level operations—such as precise clicking, dragging, and region selection—thereby constraining the reliability and scalability of end-to-end execution~\citep{cheng2024seeclick, gou2025navigating}. 

Recent studies have employed supervised fine-tuning (SFT) and reinforcement learning (RL) to train these models~\citep{gou2025navigating,qin2025ui,kang2025guirlvgincentivizeguivisual,kang2024actressactiveretrainingsemisupervised,kang2024visualgroundingattentiondrivenconstraint}; however, their performance remains suboptimal on complex or high-resolution user interfaces. For example, on the high-resolution GUI visual grounding benchmark ScreenSpot-Pro~\citep{li2025screenspot}, recently released $7$B models achieve only around $50\%$ accuracy~\citep{yang2025gta1,gu2025ui,tang2025sea}, which is not practical.

% Graphical‑user‑interface (GUI) agents powered by multimodal large language models (MLLMs) are rapidly reshaping how software is tested, automated, and made accessible. By perceiving both pixels and text, these systems can execute end‑to‑end tasks that once required handcrafted scripts—ranging from cross‑platform application testing to general desktop assistance. \textcolor{red}{Why GUI-Agent is important?}

% Yet the practical impact of GUI agents is throttled by a persistent bottleneck: reliable visual grounding—the ability to map textual commands (e.g., “click the blue Send button under Settings”) to the exact pixel region of the intended element. \textcolor{red}{To build GUI-Agent, a bottleneck: gui-visual grouding}

% Although recent studies have applied supervised fine‑tuning (SFT) and reinforcement learning (RL) to further train these models, their visual‑grounding performance on high‑resolution screens remains far from satisfactory.\textcolor{red}{recently work, why it is not good}

\begin{figure}[htbp]
  \centering
  \includegraphics[width=0.99\textwidth]{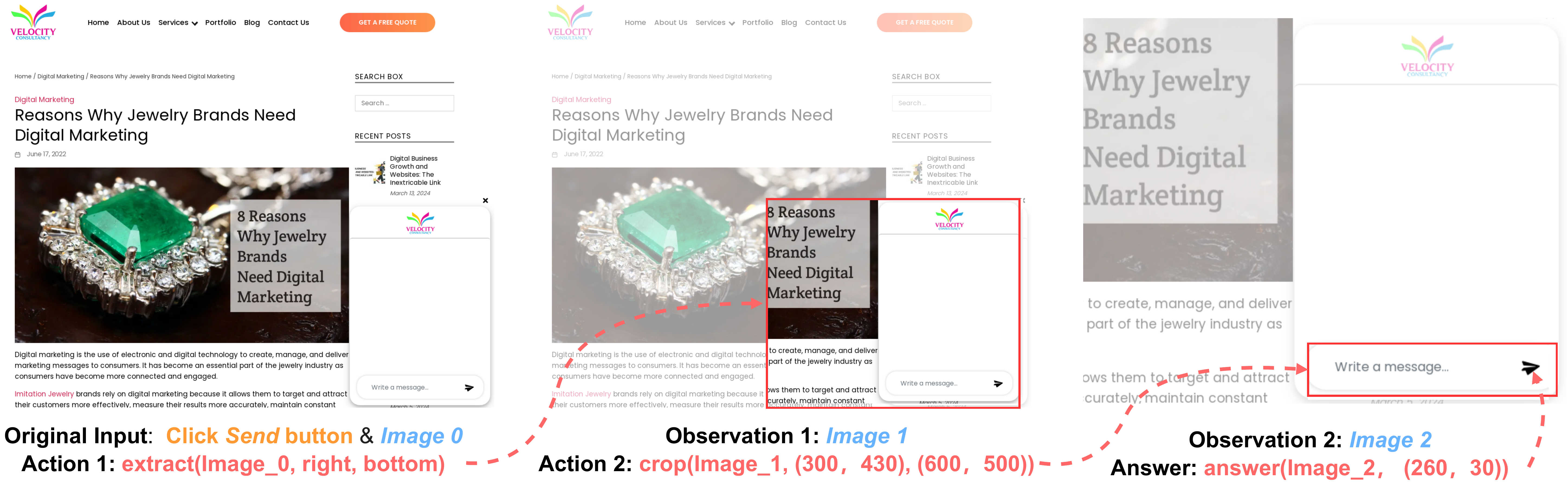}
  \caption{GUI-Spotlight pipeline. \textcolor{Orange}{Orange text} denotes the user’s original input; \textcolor{LightSkyBlue}{blue text} indicates the image provided in each dialogue turn; \textcolor{Red}{red text} indicates the command generated by the model in that turn. Red boxes highlight the newly cropped images produced by the model’s command.}
  \label{fig:pipeline}
\end{figure}

To overcome this limitation, we propose \textsc{GUI-Spotlight}, a novel GUI visual grounding model that \textit{thinks with the image} and dynamically narrows its focus like a spotlight, iteratively homing in on the target for more precise on-screen actions. To achieve this, \textsc{GUI‑Spotlight} is equipped with a set of specialized visual tools--—\textit{crop}, \textit{extract}, and \textit{find color}—--that allow it to iteratively interrogate sub-regions of the screen and progressively refine its search until the target is pinpointed with high precision. As shown in Fig.~\ref{fig:pipeline}, given the user’s original instruction \textit{Click the Send button} and the original screenshot \textit{Image 0}, GUI-Spotlight iteratively invokes tools to progressively narrow its focus to the precise click location. After each invocation, the newly cropped image is appended to the dialogue history. The model returns the final answer once coordinate confidence is sufficient.

\textsc{GUI‑Spotlight} is trained in three stages. In Stage $1$, we collect multi-turn tool-usage dialogues and warm up the model via SFT. In Stages $2$ and $3$, we conduct reinforcement learning with a modified Group Sequence Policy Optimization (GSPO) algorithm~\citep{zheng2025group}, enabling the model to learn when and how to use tools effectively, yielding a robust policy that continually improves and substantially boosts visual grounding accuracy. Details are provided in Section~\ref{sec:training}.
% We first employ supervised fine-tuning to teach the agent how to use each tool. Subsequently, we leverage the Group Relative Policy Optimization (GRPO) algorithm, enabling the agent to learn \textit{when} and \textit{how} to combine these tools effectively, creating a robust policy that improves continuously and substantially boosts visual grounding accuracy.

Our contributions are as follows:
\begin{enumerate}
  \item We introduce \textsc{GUI-Spotlight}, a think-with-image visual grounding model that performs iterative spotlighting with tool coordination; it achieves $\mathbf{52.8\%}$ accuracy on \textsc{ScreenSpot-Pro} and $\mathbf{23.4\%}$ on UI-Vision, substantially outperforming comparable $7$B baselines.
  \item We modify GSPO for multi-tool reinforcement learning, yielding a stable training procedure that improves sample efficiency and final grounding accuracy.
  \item We document and consolidate our attempts—including negative results—on algorithms, reward design, and training settings, providing practical insights for agentic visual grounding models with coordinated tool use.
\end{enumerate}

\section{Related Work}
\paragraph{GUI grounding.}
Recent progress in GUI agents has been propelled by specialised GUI–grounding models that map natural-language references to screen coordinates~\citep{you2024ferret}. SeeClick~\citep{cheng2024seeclick} introduced the task via ScreenSpot, showing that grounding pre-training improves end-to-end success across mobile, web, and desktop UIs. UGround~\citep{gou2025navigating} scaled data collection to a universal, cross-platform backbone, while OS-Atlas~\citep{wu2024atlas} expanded synthetic coverage and improved out-of-distribution transfer. UI-TARS~\citep{qin2025ui} explored capacity scaling with native mouse-and-keyboard action spaces, and Aguvis~\citep{sarch2025grounded} proposed a modular “ground-then-plan” pipeline for fully open-source autonomy. Together, these works highlight data scale, model capacity, and modular design as complementary levers, yet accuracy on dense, high-resolution, cluttered interfaces remains challenging.

\paragraph{Reinforcement learning for grounding.}
RL offers a complementary path by casting localization as sequential decision-making. UniVGR1~\citep{bai2025univg} iteratively refines boxes and attains state-of-the-art results on RefCOCO-style corpora; \textsc{Grounded RL for Visual Reasoning} ties rewards to localization correctness for multi-step reasoning~\citep{sarch2025grounded}. In GUI automation, self-evolutionary RL reduces off-target actions without extra labels~\citep{yuan2025enhancing}. \textsc{Ground-R1}~\citep{cao2025ground} further shapes rewards to balance overlap, textual relevance, and action efficiency, improving generalization to unseen categories and layouts. Despite these advances, high-resolution or heavily cluttered screens remain difficult—motivating our focus-refinement approach.

\section{Method}

\subsection{Agentic Interaction Framework}
\noindent\textbf{Inference Pipeline} As shown in Algorithm~\ref{alg:model_inference}, Given a text description \(d\) and the original image \(I_0\), we maintain a registry
\(R=\{\, i \mapsto (I_i,\boldsymbol{\delta}_i)\,\}\),
where \(\boldsymbol{\delta}_i\) denotes the top-left offset of \(I_i\) w.r.t.\ \(I_0\),
and a message history \(\mathcal{H}\) initialized with \((d, I_0)\).
At round \(t\), we send \(\mathcal{H}\) to the model, which returns either
\(\textit{Action}(i,\mathrm{Tool},\mathrm{args})\) or
\(\textit{Stop}(i,(x_{\mathrm{rel}},y_{\mathrm{rel}}))\).
For \(\textit{Action}\), executing the tool on \(I_i\) yields a new image \(I_{i+1}\), an information message describing the image and
the offset $\boldsymbol{\delta}_{i+1}$ of image \(I_{i+1}\) relative to the original image.
Register $I_j,\boldsymbol{\delta}_{i+1}$ in $R$, and append the result to \(\mathcal{H}\).
When \(\textit{Stop}\) is returned, the absolute coordinate on \(I_0\) is calculated and returned.
\begin{algorithm}[htbp]
\footnotesize
\caption{GUI-Spotlight Inference Pipeline}
\label{alg:model_inference}
\KwIn{Text description $d$ of the target element; original image $I_0$.}
\KwOut{Absolute coordinate of the element $(x_{\mathrm{abs}}, y_{\mathrm{abs}})$ on $I_0$ or None.}

\textbf{Registry}~$R = \{\, i \mapsto (I_i,\ \boldsymbol{\delta}_i) \mid i \in \mathbb{N} \,\}$, 
where $I_i$ is the $i$-th image, $\boldsymbol{\delta}_i=(\delta_i^x,\delta_i^y)$ is the top-left offset w.r.t.\ $I_0$, and $I_0$ is the original image.\;
\textbf{Initialization}: assign $\boldsymbol{\delta}_0=(0,0)$ to $I_0$; Message History $\mathcal{H} \gets \{(d,\, I_0)\}$.\;

\For{$t=1$ \KwTo $T_{\max}$}{
  $\textit{Action} \gets \mathrm{Model}(\mathcal{H})$\;

  \uIf{$\textit{Action} = \mathrm{Stop}(i,\,(x_{\mathrm{rel}}, y_{\mathrm{rel}}))$}{
    $(x_{\mathrm{abs}}, y_{\mathrm{abs}}) \leftarrow R[i].\boldsymbol{\delta}_i + (x_{\mathrm{rel}}, y_{\mathrm{rel}})$\;
    \Return $(x_{\mathrm{abs}}, y_{\mathrm{abs}})$\;
  }
  \ElseIf{$\textit{Action} = \mathrm{Tool}(i,\,\mathrm{args})$}{
    $(I_{i+1},\, \mathrm{info},\, \boldsymbol{\delta}_{i+1}) \leftarrow \mathrm{Tool}(R[i].I_i,\, \mathrm{args})$\;
    $R[i+1] \leftarrow (I_{i+1},\, \boldsymbol{\delta}_{i+1})$\;
    $\mathcal{H} \gets \mathcal{H} \cup \{(I_{i+1},\, \mathrm{info})\}$\;
  }
}
\Return None\;
\end{algorithm}

\noindent\textbf{Tool Functions.} We design three visual grounding tools; their functionality, inputs, and outputs are summarized in Table~\ref{tab:spotlight-tools-4col}. \texttt{extract}: quadrant crop by position for coarse focus narrowing. \texttt{find\_color}: color-guided focusing—
slides a window to locate the region of closest color match to a target RGB by minimizing the perceptual color difference (\(\Delta E\), the Euclidean distance in CIE Lab space), then extracts a centered crop.
% slide a window to locate the position with minimal \(\Delta E\) to the target RGB in CIE Lab, then center a window there. \(\Delta E\) denotes the perceptual color difference in CIE Lab (Euclidean distance); smaller is more similar. 
\texttt{crop}: rectangular crop specified by opposite corners for fine-grained focus. All tools return the cropped image, an information message, and the top-left offset of the crop relative to the original image.

\begin{table}[htbp]
\centering
\footnotesize
\caption{Tool functions used in GUI-Spotlight.}
\label{tab:spotlight-tools-4col}
\begin{adjustbox}{width=\linewidth} % or width=\textwidth
\begin{tabular}{@{}C{.14\linewidth} C{.36\linewidth} C{.22\linewidth} C{.28\linewidth}@{}}
\toprule
\textbf{Function} & \textbf{Function Logic} & \textbf{Input} & \textbf{Output} \\
\midrule
\texttt{extract} &
Quarter crop (\(\tfrac{1}{2}W \times \tfrac{1}{2}H\)) by position; validate options; enforce minimum crop size; compute top-left offset. &
\texttt{Image}; \texttt{x\_pos} \(\in\)\{left, center, right\}; \texttt{y\_pos} \(\in\)\{top, center, bottom\}. &
\texttt{Image}; \texttt{Info}; \texttt{offset} \((\Delta{x},\Delta{y})\) or \texttt{None}. \\
\midrule
\texttt{find\_color} &
Scan \(10{\times}10\) patches (stride 10); pick minimal \(\Delta E\) (CIE Lab) to target RGB; center a \(ws{\times}ws\) window. &
\texttt{Image}; \texttt{target\_rgb} \(=(r,g,b)\). &
\texttt{Image}; \texttt{Info}; \texttt{offset} \((\Delta{x},\Delta{y})\) or \texttt{None}. \\
\midrule
\texttt{crop} &
Rectangular crop with bounds/order/min-size checks; optional \(\pm1\)px adjustment for edge case; compute offset. &
\texttt{Image}; \texttt{top\_left} \(=(x_1,y_1)\); \texttt{bottom\_right} \(=(x_2,y_2)\). &
\texttt{Image}; \texttt{Info}; \texttt{offset} \((\Delta{x},\Delta{y})\) or \texttt{None}. \\
\bottomrule
\end{tabular}
\end{adjustbox}
\end{table}

\subsection{Training}
\label{sec:training}
In this section, we present the datasets we collected for training, the three-stage training pipeline, and the reward formulation used during reinforcement learning.
\subsubsection{Dataset} 

\noindent\textbf{\textit{Data Collection}} During our investigation, we observed that many open-source datasets already exist at low resolution. To provide more challenging tasks during training and thereby encourage deeper reasoning, we collected an additional $15$K high-resolution samples. Specifically, we employ a Selenium-based headless browser to batch-load webpages at a fixed resolution and automatically detect common interactive elements. For visible elements of reasonable size, we extract their readable text, crop element-level images from full-page screenshots, and store the text together with bounding-box metadata, organized per site. This pipeline enables large-scale collection and cleaning, resulting in a consistent dataset of clickable components for downstream training and evaluation.

\noindent\textbf{\textit{Data Cleaning}} Raw instruction-following datasets often contain significant noise, such as blurry screenshots, ambiguous instructions, or inaccurate annotations. To ensure the quality of our training data, we developed a rigorous filtering pipeline to curate a high-fidelity dataset. First, we perform image-level pre-filtering, discarding images if their Laplacian variance is below $100.0$ (Clarity) or if the ground-truth bounding box covers less than $1\%$ of the image area (Visibility).

The core of the filtering process uses the \texttt{Qwen2.5-VL-72B} model to audit each instruction-response pair via the following three evaluations:
\begin{itemize}[nosep]
    \item \textbf{Instruction Quality (IQ):} To filter out unclear instructions, the model rates clarity and uniqueness on a $0$–$10$ scale and filters out ambiguity by accepting only those scoring $\ge 6$.
    \item \textbf{Bounding Box Accuracy (BA):} To verify label accuracy, the model's prediction ($B_p$) is compared against the ground-truth ($B_{gt}$). The resulting accuracy score, also on a 0-10 scale, must be $\ge 6$, as determined by the formula $S_{BA} = 5 \cdot \frac{|B_p \cap B_{gt}|}{|B_{gt}|} + 5 \cdot \frac{|B_p \cap B_{gt}|}{|B_p|}$.
    \item \textbf{Consistency (CON):} To ensure the model's interpretation is stable and not coincidental, this self-verification step requires the Intersection over Union (IoU) between two independently generated boxes ($B_{IQ}, B_{BA}$) to be at least 0.40, as calculated by $IoU = \frac{|B_{IQ} \cap B_{BA}|}{|B_{IQ} \cup B_{BA}|}$.
\end{itemize}
A sample is retained only if it passes all three filters. We applied this pipeline to both the public UGround dataset~\cite{gou2025navigating} and our newly collected high-resolution data. On the UGround dataset, the process retained approximately $50\%$ of the data as a high-quality subset. For our high-resolution dataset, we enhanced the pipeline with additional functions to filter for websites in major languages and recognizable interfaces, yielding a refined dataset of $11.6K$ samples. This comprehensive cleaning ensures a consistent standard of quality across both datasets used in our work.

% \noindent\textbf{\textit{Data Cleaning}} We clean both the UGround dataset~\cite{gou2024navigating} and our high-resolution in-house data with a three-step perception–localization–consistency pipeline. First, we apply visibility and clarity filters (the candidate bbox must lie within the screen with a minimum area, and image sharpness must exceed a Laplacian-variance threshold). Next, the \texttt{Qwen2.5-VL-72B} performs dual localization per instruction: an Instruction Quality pass that verifies executability and yields a confidence score, and a Box Accuracy pass that outputs the element’s bounding box. Finally, we compute consistency via the IoU between the two predicted boxes and retain samples only if joint thresholds on Instruction Quality confidence, Box Accuracy , and consistency are met. This design jointly enforces executability, localization accuracy, and cross-check consistency, substantially reducing noise and non-parseable cases.

\subsubsection{Three stages training}

\textsc{GUI-Spotlight} training was carried out in three distinct stages. The training pipeline and algorithms are presented below, and the full hyperparameter settings are provided in the Appendix~\ref{apd:3_stages_training}. 

\noindent\textbf{\textit{Stage 1}} We first executed the same inference pipeline with \texttt{Qwen2.5-VL-72B}~\cite{bai2025qwen2} on the filtered UGround dataset and collected $2561$ multi-turn dialogue trajectories with tool invocations. We then used these trajectories to warm up the initial models via supervised imitation, starting from \texttt{UI-TARS-1.5-7B}~\cite{qin2025ui} and \texttt{Qwen2.5-VL-7B-Instruct}~\cite{bai2025qwen2}. This stage teaches the models to compose multiple tools, providing a solid initialization for subsequent RL.

\noindent\textbf{\textit{Stage 2}}
We further optimize the model via reinforcement learning using $12$K samples from the filtered UGround dataset, and we modify the original GSPO~\cite{zheng2025group} objective as follows:
% \begin{align}
% \mathcal{L}_{\mathrm{ours}}(\theta)
% ~=~ -\mathcal{J}_{\mathrm{GSPO}}(\theta)
% \;+\; \bm{\lambda\,{\mathcal{L}^{\prime}(\theta)}}
% \end{align}
{\footnotesize
\begin{align}
\mathcal{J}_{\mathrm{Ours}}(\theta)
&= \mathbb{E}_{x\sim\mathcal{D},\,\{y_i\}_{i=1}^G \sim \pi_{\theta_{\mathrm{old}}}(\cdot\,|\,x)}
\left[
\frac{1}{G}\sum_{i=1}^{G}
\min\!\Big(
s_i(\theta)\,\widehat{A}_i,\;
\mathrm{clip}\big(s_i(\theta),\,1-\varepsilon,\,1+\varepsilon\big)\,\widehat{A}_i
\Big)
\right] \nonumber + \textcolor{red}{\lambda \,\mathcal{J}^{\prime}(\theta)}
\end{align}}

where 
\begin{align}
\nonumber
\widehat{A}_i
&= \frac{ r(x,y_i)- \mathrm{mean}\big(\{r(x,y_j)\}_{j=1}^{G}\big) }
         { \mathrm{std}\big(\{r(x,y_j)\}_{j=1}^{G}\big) }, & s_i(\theta)= \exp\!\left(
\frac{1}{|y_i|}
\sum_{t=1}^{|y_i|}
\log \frac{\pi_{\theta}(y_{i,t}\,|\,x, y_{i,<t})}
          {\pi_{\theta_{\mathrm{old}}}(y_{i,t}\,|\,x, y_{i,<t})}
\right).
\end{align}
{\footnotesize
\begin{align}
\nonumber
\mathcal{J}^{\prime}(\theta) = \frac{1}
      {\displaystyle\sum_{b=1}^{B}\sum_{t=1}^{L_b} C_b\,M_{b,t}\;+\;\varepsilon}\displaystyle\sum_{b=1}^{B}\sum_{t=1}^{L_b}
      C_b\,M_{b,t}\,\log \pi_{\theta}\!\big(y_{b,t}\mid s_{b,t}\big)
\end{align}}
$B$ is the batch size, $L_b$ is the sequence of sample $b$, $\lambda>0$ is a mixing weight; in stage~$2$ we set $\lambda=1$; and
$\bm{\mathcal{J}^\prime(\theta)}$ is computed by first filtering to the subset
of samples whose outputs are both format-valid and result-correct, and then
averaging the token-level cross-entropy over this subset. $M_{b,t}\in\{0,1\}$ is the completion mask. And 
$C_b\in\{0,1\}$ is the sample mask for result-correct and format-correct cases.
$\varepsilon>0$: a tiny constant to avoid division by zero.

This auxiliary term $\bm{\mathcal{J}^{\prime}(\theta)}$ is introduced to stabilize reinforcement learning in multi-turn tool-use scenarios. In its absence, broad autonomous exploration often leads the model to generate non-parseable tool formats, resulting in sparse and volatile rewards. Such instability induces high-variance gradients and excessively large policy updates, which in turn cause parameter drift and ultimately lead to training collapse. A detailed analysis is provided in Section~\ref{sec:RL_selection}.

\noindent\textbf{\textit{Stage 3}} 
Using $4000$ high-resolution samples that we collected, we further refined training once the tool-call format had stabilized. Specifically, we reduced the mixing weight to $\lambda=0.01$ and revised the sample mask $C_b$ in $\bm{\mathcal{J}^{\prime}(\theta)}$: rather than retaining all results or only format-correct cases, we applied bucketed uniform sampling across tool types.

Bucket construction:
\begin{align}
S_t&=\{\,b:\ \mathrm{tool}(b)=t,\ \mathrm{correct}(b)=1,\ \mathrm{format}(b)=1\,\},
n_{\min}=\min_{t\in\mathcal{T}}|S_t|
,
\hat S_t\subseteq S_t,\ |\hat S_t|=n_{\min}.\nonumber
\end{align}
Sample mask:
{\footnotesize
\begin{align}
\nonumber
C_b :=
\begin{cases}
1, & b\in \bigcup_{t\in\mathcal{T}}\hat S_t,\\
0, & \text{otherwise.}
\end{cases}
\end{align}}
Where $t\in\mathcal{T}$ is the tool type; $b$ is the sample index. $S_t$ is bucket for tool $t$.

\begin{wrapfigure}{r}{0.5\textwidth} % r=
  \centering
  \vspace{-14pt}
\includegraphics[width=0.98\linewidth]{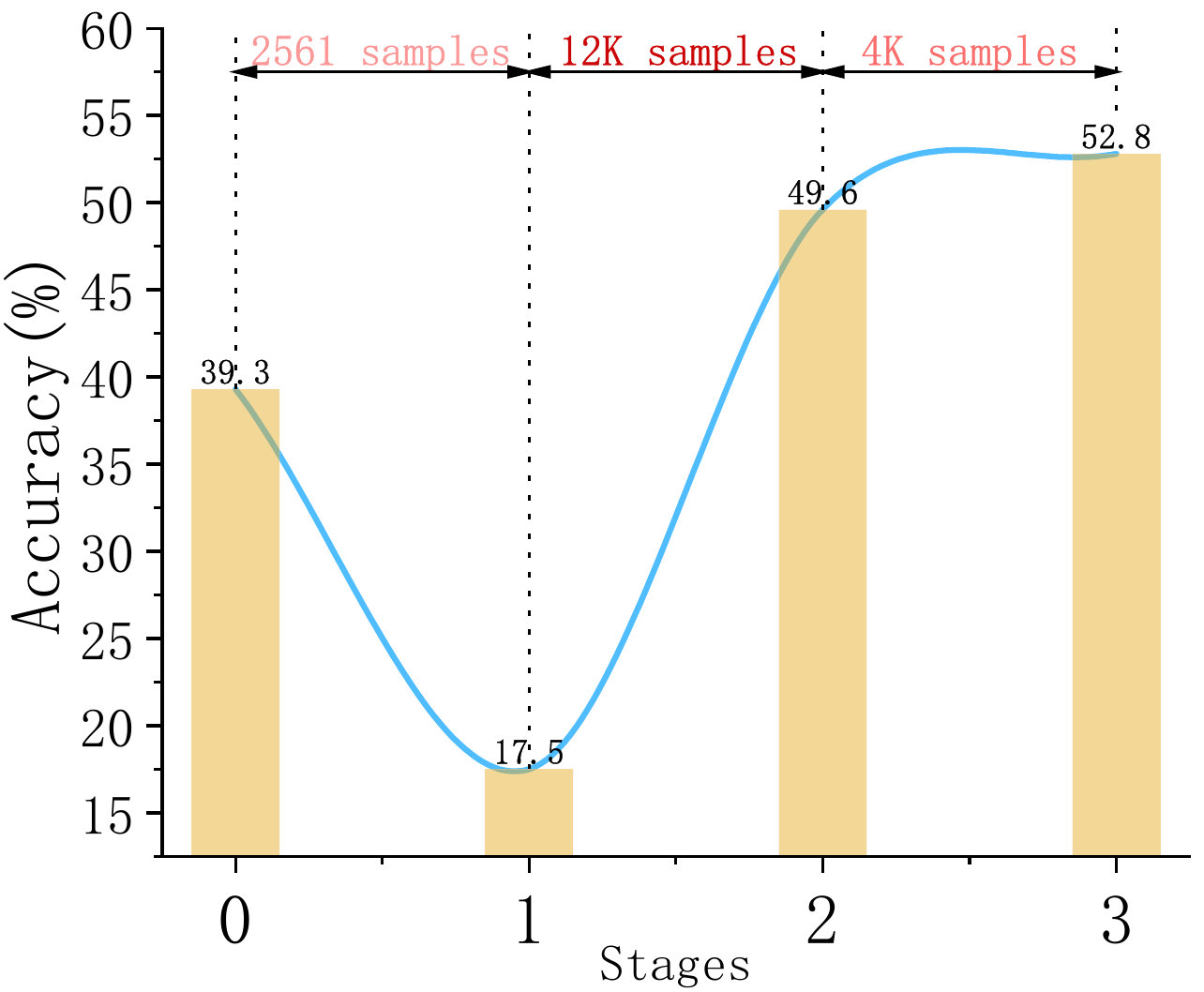}
  \caption{ScreenSpot-Pro accuracy over training.}
  \label{fig:3_stage_training}
  \vspace{-6pt}
\end{wrapfigure}

The evolution of test accuracy on the ScreenSpot-Pro benchmark across the three training stages is shown in Figure~\ref{fig:3_stage_training}. Using \texttt{UI-TARS-1.5-7B} as an base model:
Stage $1$: We perform one epoch of SFT on $2561$ trajectories. After warm-up, the model learns to invoke multiple tools but remains under-aligned.
Stage $2$: We train on $12$K examples with RL, yielding a substantial accuracy gain.
Stage $3$: we then introduce additional $4000$ samples to encourage exploration, leading to a further improvement in accuracy.

% \begin{wrapfigure}{r}{0.65\textwidth} % r=
%   \centering
%   % \vspace{-6pt}
% \includegraphics[width=0.98\linewidth]{Figures/3_stages_training.pdf}
%   \caption{Accuracy trajectory over three training stages.}
%   \label{fig:3_stage_training}
%   % \vspace{-6pt}
% \end{wrapfigure}

%------------------------------------------------
\subsubsection{Reward Design}
\label{subsec:reward}
%------------------------------------------------
We combine five rewards into a weighted sum
{\footnotesize\[
R=\sum_{k=1}^{5}\alpha_k r_k,\
(\alpha_1,\dots,\alpha_5)=(0.30,0.25,0.05,0.20,0.20),\ \sum_k \alpha_k=1.
\]}
They are summarized in Table~\ref{tab:reward function}.
$r_1$ \texttt{Answer} is a sparse reward for a correct final answer.
$r_2$ \texttt{Crop} uses IoU to provide dense reward toward the ground-truth region.
$r_3$ \texttt{Extract} and $r_4$ \texttt{Find\_Color} supply binary intermediate feedback for quadrant/color-guided focusing.
$r_5$ \texttt{Format} checks the syntactic validity of tool calls to improve the stability of training.
\begin{table}[htbp]
\centering
\caption{Reward components used for RL training. $B^\star$ denotes the ground-truth bounding box.
}
\label{tab:reward function}
\scriptsize
\resizebox{\linewidth}{!}{%
\begin{tabular}{@{}ccccc@{}}
\toprule
\textbf{\#} & \textbf{Name} & \textbf{Definition} & \textbf{Type} & \textbf{Weight} \\
\midrule
$r_1$ & \texttt{Answer} &
$1$ if the final answer 
the predicted coordination $(\hat{x},\hat{y})$ lies inside the $B^\star$ $(x_1,y_1,x_2,y_2)$; $0$ otherwise. &
\textit{sparse} & $0.3$ \\[2pt]
$r_2$ & \texttt{Crop} &
For each \texttt{\textless crop\textgreater} call $i$ we compute
$\operatorname{IoU}_i=\frac{|\,\hat B_i \cap B^\star|}{|\,\hat B_i \cup B^\star|}\!\in[0,1]$.
 &
\textit{dense} & $0.25$ \\[2pt]
$r_3$ & \texttt{Extract} &
Each quadrant extracted by \texttt{\textless extract\textgreater} yields $1$ if it fully contains $B^\star$, else $0$.
The reward is their mean. &
\textit{sparse} & $0.05$ \\[2pt]
$r_4$ & \texttt{Find\_Color} &
Returns $1$ when the $200\times200$ color-match window covers $B^\star$; $0$ otherwise. &
\textit{sparse} & $0.2$ \\[2pt]
$r_5$ & \texttt{Format} &
$1$ if the assistant’s tool call is syntactically valid; $0$ otherwise. &
\textit{sparse} & $0.2$ \\
\bottomrule
\end{tabular}%
}
\vspace{-10pt}
\end{table}
The exploration of how different reward designs affect the final model performance are shown in Section~\ref{sec:reward_design}.

\section{Empirical Insights}
Throughout this study, we systematically evaluated how different algorithms and reward designs affect the model’s final performance; the experimental records are presented below.

\subsection{RL Algorithm Selection}
\label{sec:RL_selection}

\begin{figure}[t]
  \centering
  \begin{minipage}[t]{.495\textwidth}
    \centering
    \includegraphics[width=\linewidth]{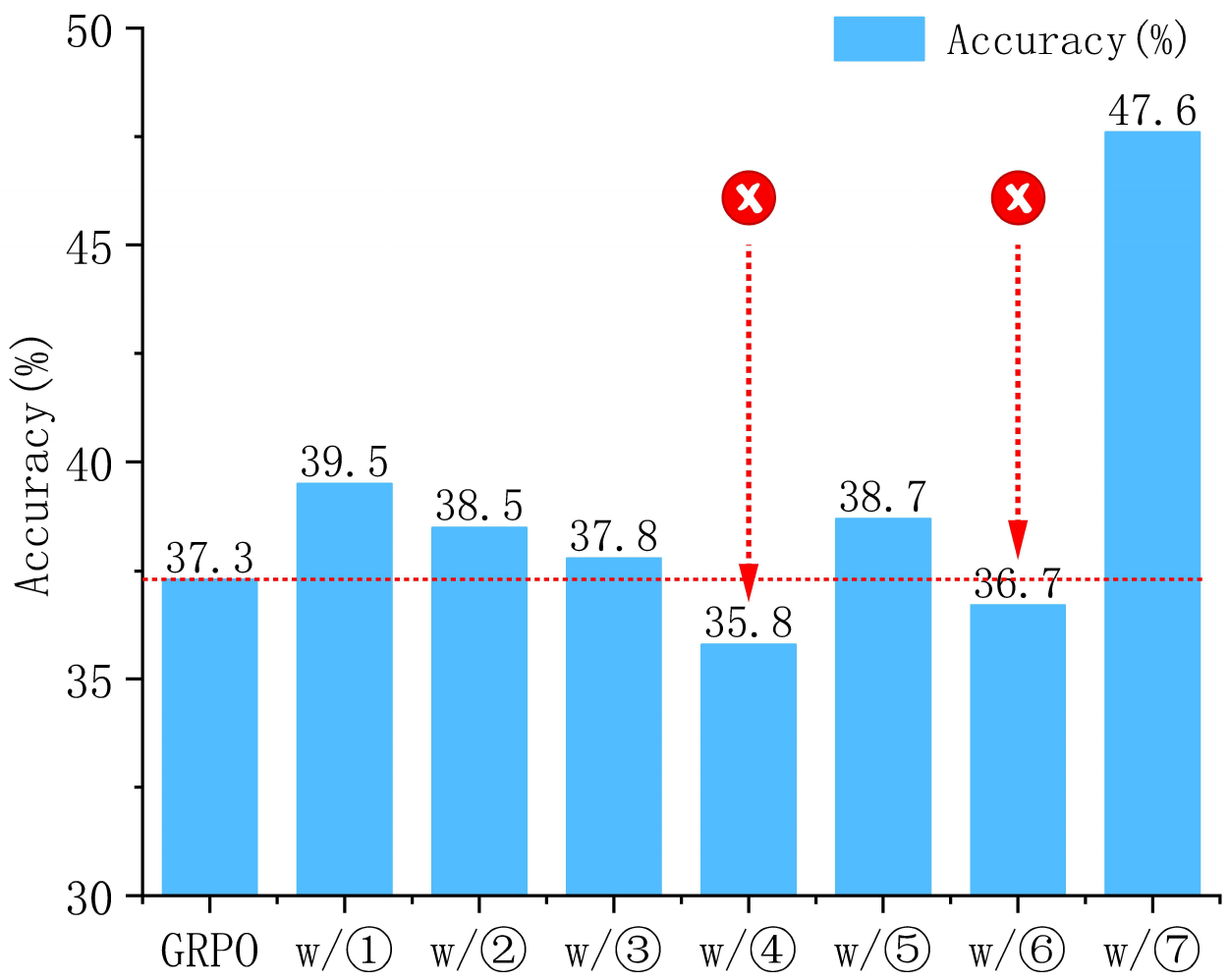}
  \end{minipage}\hfill
  \begin{minipage}[t]{.495\textwidth}
    \centering
    \includegraphics[width=\linewidth]{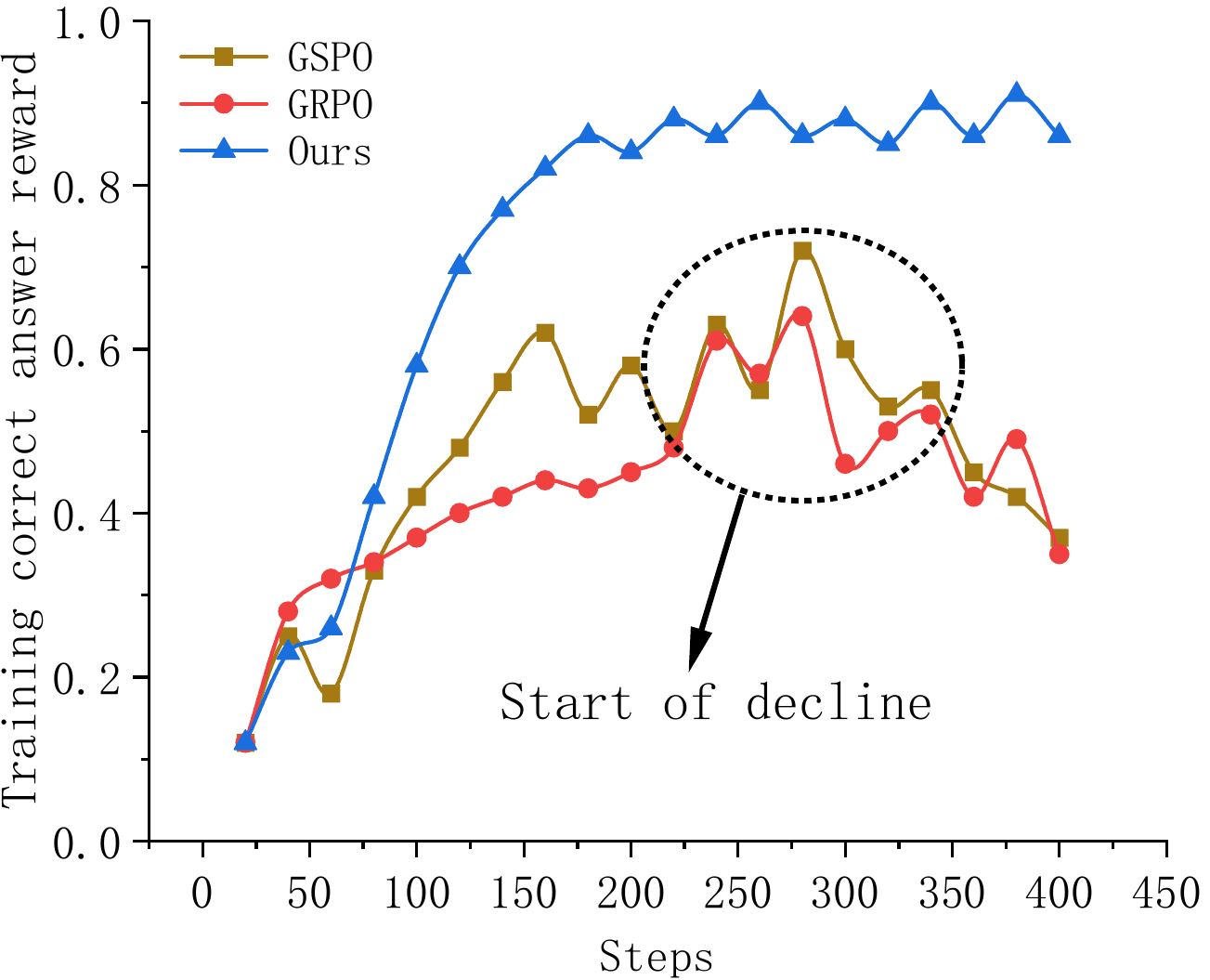}
  \end{minipage}
    \caption{\textbf{Left}: Impact of different RL variants. \textbf{Right}: A comparison of algorithm training dynamics .~\includegraphics[height=0.7em]{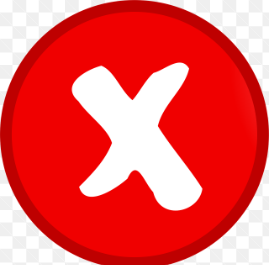}\ denotes \textit{discarded}. Items \ding{172}–\ding{178} are described in the first paragraph of Section~\ref{sec:RL_selection}.}
  \label{fig:different_rl}
  \vspace{-15pt}
\end{figure}

To investigate how different RL algorithms perform on the multi-turn, tool-using GUI visual grounding task, we benchmark a suite of GRPO-based~\cite{shao2024deepseekmath} improvements alongside our own variants. Specifically, the evaluated techniques include \ding{172} sequence-level importance-ratio sampling~\cite{zheng2025group}, \ding{173} Clip-Higher~\cite{yu2025dapo}, \ding{174} KL-term removal~\cite{yu2025dapo}, \ding{175} retaining only the top $p\%$ most-uncertain prompts~\cite{wang2025ragen}, and \ding{176} adding a positive-example LM loss~\cite{yue2025vapo}. In addition, we introduce two of our own designs: \ding{177} continuously updating the reference policy, and \ding{178} tool-filtered positives with an additional cross-entropy loss. To isolate the effect of the RL objective, we first conduct a Stage-1 warm-up and then compare RL algorithms under identical settings. Concretely, we initialize from the same SFT checkpoint of \texttt{UI-TARS-1.5} trained for one epoch on $2561$ multi-turn tool-invocation trajectories, run $400$ RL steps for each method, and evaluate on the \textsc{ScreenSpot-Pro} benchmark; training parameters are provided in the Appendix~\ref{apd:alg_exploration}. As shown in the left panel of Figure~\ref{fig:different_rl}. In our setting, selecting only the highest $p\%$ most uncertain prompts and continuously updating the reference policy both degrade accuracy. Therefore, we discard these two modifications and keep the remaining improvements.

Moreover, as shown in the right panel of Figure~\ref{fig:different_rl}, we observe that the tool-filtered positives with an additional cross-entropy loss effectively prevents RL collapse on this task. Vanilla GRPO or GSPO begins to oscillate around 300 steps, with outputs increasingly violating the tool-call syntax, leading to a gradual drop in accuracy. In contrast, once we add the additional cross-entropy loss , the training curve no longer degrades and instead continues to improve.

\subsection{Reward Design}
\label{sec:reward_design}
To investigate how different reward formulations affect model performance on the multi-turn, tool-using GUI visual grounding task, we conducted experiments using the same settings as Section~\ref{sec:RL_selection}.

First, We study how the different types of  \texttt{Answer} reward affects the final performance. Specifically, we compare two formulations:
\begin{enumerate}[nosep]
  \item \textbf{Binary sparse reward}: assign $r_{\text{answer}}=1$ if the predicted click $(x,y)$ lies inside the ground-truth box $(x_1,y_1,x_2,y_2)$; otherwise $r_{\text{answer}}=0$.
  \item \textbf{Center-shaped dense reward}: if $(x,y)$ is inside the box, let $c_x=(x_1+x_2)/{2}$ and $c_y=(y_1+y_2)/{2}$ be the box center,
  and $(x_2-x_1)/2$, $(y_2-y_1)/2$ the half-width/half-height. Define the normalized Chebyshev distance
  {\footnotesize\[
  d=\max\!\Bigl(\frac{|x-c_x|}{(x_2-x_1)/2},\; \frac{|y-c_y|}{(y_2-y_1)/2}\Bigr)\in[0,1],\quad
  \text{closeness}=1-d,
  \]}
  and set
  {\footnotesize\[
  r_{\text{answer}} = 1 + \text{bonus}_{\max}\cdot (\text{closeness})^{\gamma}(\text{if inside}), 
   \quad r_{\text{answer}} = 0\quad(\text{otherwise}),
  \]}
  where $\gamma\!\ge\!1$ shapes the curvature around the center and $\text{bonus}_{\max}$ controls the maximum extra credit at the center.
\end{enumerate}

\begin{figure}[t]
  \centering
  \begin{minipage}[t]{.49\textwidth}
    \centering
    \includegraphics[width=\linewidth]{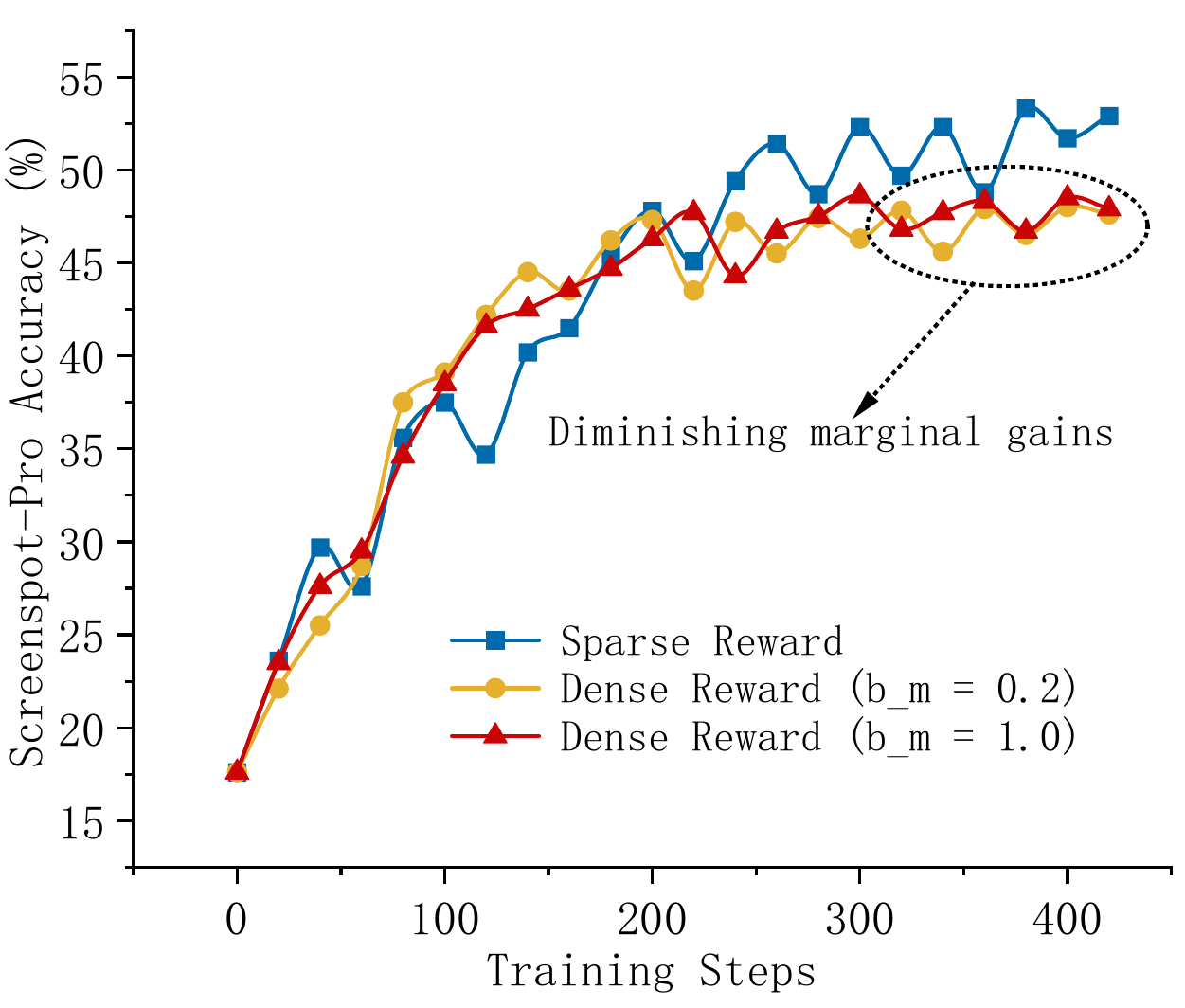}
  \end{minipage}\hfill
  \begin{minipage}[t]{.49\textwidth}
    \centering
    \includegraphics[width=\linewidth]{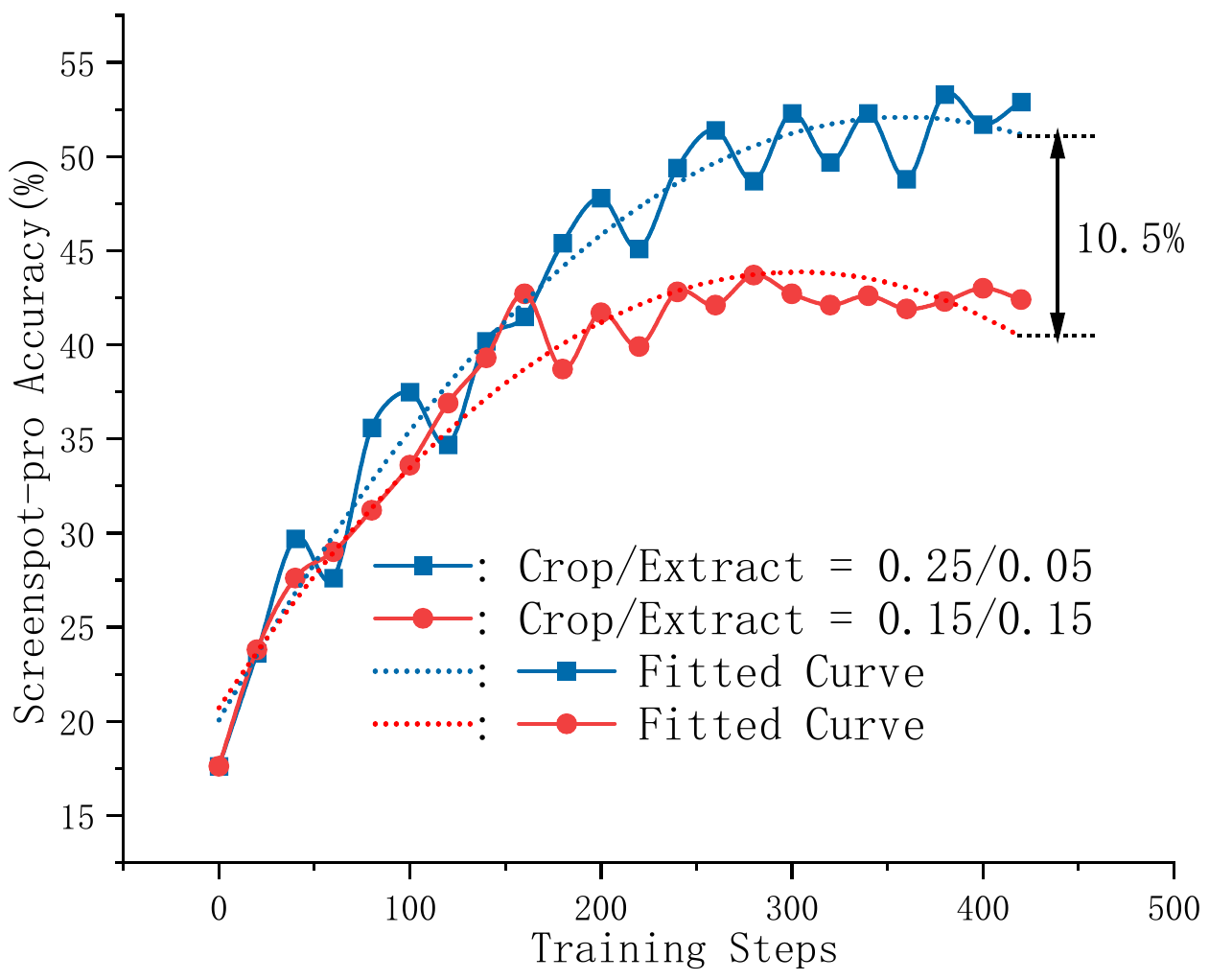}
  \end{minipage}
  \vspace{-10pt}
  \caption{\textbf{Left}: Comparison of dense and sparse \texttt{Answer} rewards. \textbf{Right}: Comparison of different \texttt{Crop/Extract} reward ratios. b\_m: $\text{bonus}_{\max}$}
  \label{fig:RL_comparision}
  \vspace{-10pt}
\end{figure}

Based on the experimental results shown in the left panel of Fig.~\ref{fig:RL_comparision}, employing a dense  \texttt{Answer} reward results in marginally lower post-convergence accuracy compared to a sparse  \texttt{Answer} reward.

We next examine how the relative weighting between the \texttt{Crop} and \texttt{Extract} rewards affects final performance. As shown in the right panel of Figure~\ref{fig:RL_comparision}, moderately increasing the weight of the \texttt{Extract} reward relative to the \texttt{Crop} reward yields a substantial gain in accuracy. We attribute this to the fact that \texttt{Extract} is easier to use than \texttt{Crop}: it only requires indicating the approximate location of the target element, without specifying precise coordinates for a bounding box.

\vspace{-10pt}
\section{Experiment}
\vspace{-10pt}
To evaluate the visual grounding capability of GUI-Spotlight, we benchmark it on ScreenSpot-Pro, OSWorld-G, and UI-Vision. The hardware and hyperparameters for training and evaluation, as well as the prompts, are provided in the Appendix~\ref{apd:Experiments}.
\begin{table}[ht]
  \centering
  \caption{ScreenSpot-Pro evaluation results. Besides \textsc{GUI-Spotlight}, we evaluated UI-TARS-1.5-7B using its official GitHub instructions~\cite{ui-tars-github}; results for the other models are taken from the ScreenSpot-Pro leaderboard~\cite{grounding-leaderboard}. ~\includegraphics[height=0.7em]{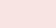}~/~\includegraphics[height=0.7em]{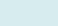}: GUI-Spotlight vs. its base.} 
  \label{tab:sspro}
\resizebox{\textwidth}{!}{%
\begin{tabular}{ccccccccc}
\hline
Model                                   & \multicolumn{1}{l}{Training Data Size} & Development & Creative & CAD    & Scientific & Office & \begin{tabular}[c]{@{}c@{}}Operating \\ System\end{tabular} & \begin{tabular}[c]{@{}c@{}}Overall \\ Average\end{tabular} \\ \hline
\rowcolor[HTML]{EFEFEF} 
\multicolumn{9}{c}{\cellcolor[HTML]{EFEFEF}\textit{\textbf{Closed-source Models}}}                                                                                                                                                                                  \\ \hline
GPT-4o                                  & -                                      & $0.7$       & $0.6$    & $1.5$  & $1.2$      & $0.9$  & $0.0$                                                       & $0.8$                                                      \\
% \multicolumn{1}{l}{Claude Computer Use}
Claude Computer Use & -                                      & $12.6$      & $16.8$   & $11.9$ & $25.8$     & $26.9$ & $8.1$                                                       & $17.1$                                                     \\
UI-TARS-1.5                             & -                                      & -           & -        & -      & -          & -      & -                                                           & $61.6$                                                     \\
Seed1.5-VL                              & -                                      & -           & -        & -      & -          & -      & -                                                           & $60.9$                                                     \\ \hline
\rowcolor[HTML]{EFEFEF} 
\multicolumn{9}{c}{\cellcolor[HTML]{EFEFEF}\textit{\textbf{Open-Source Models 72B Level}}}                                                                                                                                                                          \\ \hline
Qwen2-VL-72B-instruct                   & -                                      & $1.0$       & $0.6$    & $0.8$  & $2.4$      & $0.9$  & $0.5$                                                       & $1.0$                                                      \\
UGround-V1-72B                          & $10M$                                    & $31.1$      & $35.8$   & $13.8$ & $50.0$     & $51.3$ & $25.5$                                                      & $34.5$                                                     \\
UI-TARS-72B                             & -                                      & $40.8$      & $39.6$   & $17.2$ & $45.7$     & $54.8$ & $30.1$                                                      & $38.1$                                                     \\
Qwen2.5-VL-72B-Instruct                 & -                                      & $53.5$      & $44.9$   & $44.4$ & $59.1$     & $72.6$ & $49.5$                                                      & $53.3$                                                     \\
GTA-1-72B                               & $1.56M$                                  & $57.2$      & $51.0$   & $49.8$ & $63.0$     & $77.0$ & $57.1$                                                      & $58.4$                                                     \\
UI-Venus-72B                            & $107K$                                   & $59.5$      & $55.4$   & $57.5$ & $66.5$     & $77.8$ & $57.7$                                                      & $61.9$                                                     \\ \hline
\rowcolor[HTML]{EFEFEF} 
\multicolumn{9}{c}{\cellcolor[HTML]{EFEFEF}\textit{\textbf{Open-Source Models 32B Level}}}                                                                                                                                                                          \\ \hline
Qwen2.5-VL-32B-Instruct                 & -                                      & $48.8$      & $42.2$   & $31.0$ & $55.5$     & $64.3$ & $50.5$                                                      & $48.0$                                                     \\
GTA-1-32B                               & $1.56M$                                  & $56.2$      & $46.3$   & $38.7$ & $59.1$     & $72.2$ & $53.1$                                                      & $53.6$                                                     \\ \hline
\rowcolor[HTML]{EFEFEF} 
\multicolumn{9}{c}{\cellcolor[HTML]{EFEFEF}\textit{\textbf{Open-Source Models 7B Level}}}                                                                                                                                                                           \\ \hline
See-Click-7B                            & $1M$                                     & $0.3$       & $0.6$    & $1.9$  & $2.0$      & $0.9$  & $1.5$                                                       & $1.1$                                                      \\
Qwen2-VL-7B                             & -                                      & $1.3$       & $0.9$    & $0.4$  & $3.5$      & $3.0$  & $0.5$                                                       & $1.6$                                                      \\
UGround-7B                              & $10M$                                    & $14.7$      & $17.0$   & $11.1$ & $19.3$     & $27.0$ & $9.7$                                                       & $16.5$                                                     \\
Aguvis-7B                               & $4.2M$                                   & $16.1$      & $21.4$   & $13.8$ & $34.6$     & $34.3$ & $19.4$                                                      & $22.9$                                                     \\
\rowcolor[HTML]{F8E4E2} 
Qwen2.5-VL-7B-Instruct                  & -                                      & $26.1$      & $24.0$   & $13.0$ & $31.1$     & $45.2$ & $23.5$                                                      & $26.8$                                                     \\
UGround-V1-7B                           & $10M$                                    & $28.1$      & $31.7$   & $14.6$ & $39.0$     & $49.6$ & $24.5$                                                      & $31.1$                                                     \\
UI-TARS-7B                              & -                                      & $36.1$      & $32.8$   & $18.0$ & $50.0$     & $53.5$ & $24.5$                                                      & $35.7$                                                     \\
GUI-Actor-2VL-7B                        & $9.6M$                                   & $38.8$      & $40.2$   & $29.5$ & $44.5$     & $56.5$ & $36.2$                                                      & $40.7$                                                     \\
\rowcolor[HTML]{D5ECED} 
UI-TARS-1.5-7B                          & -                                     & $33.9$           & $33.7$        & $25.8$      & $47.6$          & $63.0$      & $33.7$                                                           & $38.7$                                                     \\
GUI-Actor-2.5VL-7B                      & $9.6M$                                   & $38.1$      & $41.3$   & $38.3$ & $50.8$     & $63.0$ & $38.8$                                                      & $44.6$                                                     \\ 
SE-GUI-7B                               & $3K$                                     & $44.5$      & $37.2$   & $42.1$ & $54.7$     & $70.4$ & $38.8$                                                      & $47.2$                                                     \\
GTA-1-7B                                & $1.56M$                                  & $44.5$      & $44.0$   & $44.4$ & $57.1$     & $75.2$ & $38.3$                                                      & $50.1$                                                     \\
V2P-7B                                  & $9.6M$                                   & $46.8$      & $43.1$   & $47.1$ & $56.3$     & $68.3$ & $45.4$                                                      & $50.6$                                                     \\ 
UI-Venus-7B                             & $107K$                                   & $50.2$      & $42.8$   & $51.0$ & $57.1$     & $67.8$ & $37.2$                                                      & $50.8$                                                     \\ \hline
\rowcolor[HTML]{EFEFEF} 
\multicolumn{9}{c}{\cellcolor[HTML]{EFEFEF}\textit{\textbf{Ours}}}                                                                                                                                                                                                  \\ \hline
\rowcolor[HTML]{F8E4E2} 
GUI-Spotlight (Init. Qwen2.5-VL-7B-Instruct)                                       & $18.5K$ \color{darkgreen}$\downarrow$                                 & $29.8$         & $29.1$      & $39.2$    & $39.8$        & $63.9$    & $24.5$                                                         & $38.7$      \color{darkgreen}$\uparrow$                                                  \\
\rowcolor[HTML]{D5ECED} 
GUI-Spotlight (Init. UI-TARS-1.5-7B)                                     & $18.5K$   \color{darkgreen}$\downarrow$                              & $53.3$         & $44.4$      & $51.0$    & $52.4$        & $71.3$    & $46.9$                                                         & $52.8$   \color{darkgreen}$\uparrow$                                                     \\ \hline
\end{tabular}
}
\end{table}

\subsection{High-Resolution Professional GUI Grounding}

ScreenSpot-Pro~\cite{li2025screenspot} is a benchmark for evaluating visual grounding on high-resolution screenshots of professional software, covering application domains such as creative tools, office platforms and so on. We use it to assess GUI-Spotlight’s accuracy on $4$K-resolution GUI screens.

As shown in Table~\ref{tab:sspro}, our model attains high accuracy, trains data-efficiently, and generalizes broadly.
\noindent \textbf{High accuracy.} \textsc{GUI-Spotlight} (init.\ UI-TARS-1.5-$7$B) reaches $\mathbf{52.8\%}$ on \textsc{ScreenSpot-Pro}, surpassing $7$B peers and remaining competitive with much larger models. It improves over its initialization across all six domains, indicating robustness to dense, icon-heavy, cluttered UIs.
\noindent \textbf{Data efficiency.} These results are achieved with only $18.5\mathrm{K}$ curated samples—far less than competing approaches that train on millions (e.g., UGround-V1-7B $\sim\!10\mathrm{M}$, V2P-7B $9.6\mathrm{M}$).
\noindent \textbf{Generality.} Starting from the non-UI-specific Qwen2.5-VL-7B-Instruct, \textsc{GUI-Spotlight} reaches an absolute $+\!11.9$ points over its raw baseline ($26.8\%$), showing that our RL objective and multi-tool coordination transfer beyond UI-specialized backbones and are robust to the choice of backbone.

\subsection{Desktop Application Visual Grounding}
To evaluate \textsc{GUI-Spotlight} in realistic desktop, we use UI-Vision~\cite{nayak2025ui}, which pairs diverse screenshots from $83$ applications across $6$ domains with dense referring expressions.

\begin{wraptable}[21]{r}{0.69\textwidth}
  \vspace{-0.5\baselineskip}         % 可选：微调与上一段的垂直间距
  \centering
  \vspace{-10pt}
  \caption{UI-Vision evaluation results. Besides \textsc{GUI-Spotlight}, we evaluated UI-TARS-1.5-7B using its official GitHub instructions~\cite{ui-tars-github}; results for the other models are taken from the UI-Venus paper~\cite{gu2025ui}. ~\includegraphics[height=0.7em]{Icons/pink_icon.png}~/~\includegraphics[height=0.7em]{Icons/blue_icon.png}: GUI-Spotlight vs. its base.}
  \label{tab:UI-Vision}
  \small                              % 可选：缩小字号以适配宽度
  \setlength{\tabcolsep}{6pt}         % 可选：列间距微调
  \renewcommand{\arraystretch}{1.1}   % 可选：行距微调
\begin{tabular}{ccccc}
\hline
Models                     & Basic        & Functional       & Spatial      & Average      \\ \hline
\multicolumn{5}{c}{\cellcolor[HTML]{EFEFEF}\textit{\textbf{Closed-Source Models}}}         \\ \hline
GPT-4o                     & $1.6$          & $1.5$              & $1.0$          & $1.4$          \\
Claude-3.7-Sonnet          & $9.5$          & $7.7$              & $7.6$          & $8.3$          \\ \hline
\multicolumn{5}{c}{\cellcolor[HTML]{EFEFEF}\textit{\textbf{Open-Source Models 72B level}}} \\ \hline
UI-TARS-72B                & $31.4$         & $30.5$             & $14.7$         & $25.5$         \\
UI-Venus-Ground-72B        & $45.6$         & $42.3$             & $23.7$         & $36.8$         \\ \hline
\multicolumn{5}{c}{\cellcolor[HTML]{EFEFEF}\textit{\textbf{Open-Source Models 7B level}}}  \\ \hline
\rowcolor[HTML]{F8E4E2} 
Qwen2.5-VL-7B              & $1.2$          & $0.8$              & $0.5$          & $0.9$          \\
OS-Atlas-7B                & $12.2$         & $11.2$             & $3.7$          & $9.0$          \\
UGround-V1-7B              & $15.4$         & $17.1$             & $6.3$          & $12.9$         \\
UI-TARS-7B                 & $20.1$         & $24.3$             & $8.4$          & $17.6$         \\
\rowcolor[HTML]{D5ECED}
UI-TARS-1.5-7B             & $22.9$         & $26.1$             & $6.6$         & $18.1$         \\
UI-Venus-Ground-7B         & $36.1$         & $32.8$             & $11.9$         & $26.5$         \\ \hline
\multicolumn{5}{c}{\cellcolor[HTML]{EFEFEF}\textit{\textbf{Ours}}}                         \\ \hline
\rowcolor[HTML]{F8E4E2} 
GUI-Spotlight (Qwen)                        & $11.1$          & $13.4$              & $1.2$          & $8.3$          \\
\rowcolor[HTML]{D5ECED}
GUI-Spotlight (UI-TARS)                 & $32.1$          & $30.2$              & $9.1$          &  $23.4$          \\ \hline
\end{tabular}
\end{wraptable}

UI-Vision evaluation results are shown in Table~\ref{tab:UI-Vision}, GUI-Spotlight trained from UI-TARS-1.5-7B surpassing its backbone UI-TARS-1.5-7B by $+5.3$ points and outperforming other 7B models and approaches the 72B UI-TARS-72B.
The variant initialized from Qwen2.5-VL-7B attains an absolute gain of $+7.4$ points over the raw
Qwen2.5-VL-7B baseline, evidencing transfer under a non-UI-specific backbone. Overall, these results indicate that our multi-tool RL training consistently improves 7B models and
narrows the gap to larger models on UI-Vision.

\begin{table}[ht]
  \centering
  \caption{OSWorld-G evaluation results. Besides \textsc{GUI-Spotlight}, we evaluated UI-TARS-1.5-7B using its official GitHub instructions~\cite{ui-tars-github}; results for the other models are taken from the UI-Venus Technical Report~\cite{gu2025ui}. ~\includegraphics[height=0.7em]{Icons/pink_icon.png}~/~\includegraphics[height=0.7em]{Icons/blue_icon.png}: GUI-Spotlight vs. its base.} 
  \label{tab:OSWorld-G}
\resizebox{\textwidth}{!}{
\begin{tabular}{ccccccc}
\hline
                         &                                                                           &                                                                                 &                                                                                  &                                                                                       &                           &                           \\
\multirow{-2}{*}{Models} & \multirow{-2}{*}{\begin{tabular}[c]{@{}c@{}}Text\\ Matching\end{tabular}} & \multirow{-2}{*}{\begin{tabular}[c]{@{}c@{}}Element\\ Recognition\end{tabular}} & \multirow{-2}{*}{\begin{tabular}[c]{@{}c@{}}Layout\\ Understanding\end{tabular}} & \multirow{-2}{*}{\begin{tabular}[c]{@{}c@{}}Fine-grained\\ Manipulation\end{tabular}} & \multirow{-2}{*}{Refusal} & \multirow{-2}{*}{Average} \\ \hline
\multicolumn{7}{c}{\cellcolor[HTML]{EFEFEF}\textit{\textbf{Closed-Source Models}}}                                                                                                                                                                                                                                                                                                                                        \\ \hline
Operator                 & $51.3$                                                                      & $42.4$                                                                            & $46.6$                                                                             & $31.5$                                                                                  & -                         & $40.6$                      \\
Gemini-2.5-pro           & $59.8$                                                                      & $45.5$                                                                            & $49.0$                                                                             & $33.6$                                                                                  & $38.9$                      & $45.2$                      \\
Seed1.5-VL               & $73.9$                                                                      & $66.7$                                                                            & $69.6$                                                                             & $47.0$                                                                                  & $18.5$                      & $62.9$                      \\ \hline
\multicolumn{7}{c}{\cellcolor[HTML]{EFEFEF}\textit{\textbf{Open-Source Models-72B Level}}}                                                                                                                                                                                                                                                                                                                                \\ \hline
UI-TARS-72B              & $69.4$                                                                      & $60.6$                                                                            & $62.9$                                                                             & $45.6$                                                                                  & -                         & $57.1$                     \\
Qwen2.5-VL-72B           & $52.6$                                                                      & $74.6$                                                                            & $74.7$                                                                             & $55.3$                                                                                  & -                         & $62.2$                      \\
UI-Venus-Ground-72B      & $82.1$                                                                      & $71.2$                                                                            & $70.7$                                                                             & $64.4$                                                                                  & -                         & $70.4$                      \\ \hline
\multicolumn{7}{c}{\cellcolor[HTML]{EFEFEF}\textit{\textbf{Open-Source Models-7B Level}}}                                                                                                                                                                                                                                                                                                                                 \\ \hline
OS-Atlas-7B              & $44.1$                                                                      & $29.4$                                                                            & $35.2$                                                                             & $16.8$                                                                                  & $7.4$                       & $27.7$                      \\
\rowcolor[HTML]{F8E4E2} 
Qwen2.5-VL-7B            & $45.6$                                                                      & $32.7$                                                                            & $41.9$                                                                             & $18.1$                                                                                  & -                         & $31.4$                      \\
UGround-7B               & $51.3$                                                                      & $40.3$                                                                            & $43.5$                                                                             & $24.8$                                                                                  & -                         & $36.4$                      \\
Aguvis-7B                & $55.9$                                                                      & $41.2$                                                                            & $43.9$                                                                             & $28.2$                                                                                  & -                         & $38.7$                      \\
UI-TARS-7B               & $60.2$                                                                      & $51.8$                                                                            & $54.9$                                                                             & $35.6$                                                                                  & -                         & $47.5$                      \\
Jedi-7B                  & $65.9$                                                                      & $55.5$                                                                            & $57.7$                                                                             & $46.9$                                                                                  & $7.4$                       & $54.1$                      \\
UI-Venus-Ground-7B       & $74.6$                                                                      & $60.5$                                                                            & $61.5$                                                                             & $45.5$                                                                                  & -                         & $58.8$                      \\
\rowcolor[HTML]{D5ECED}
UI-TARS-1.5-7B           & $67.3$                                                                      & $64.5$                                                                            & $65.2$                                                                             & $42.9$                                                                                  & -                         & $61.9$                      \\
GTA1-7B                  & $63.2$                                                                      & $82.1$                                                                            & $74.2$                                                                             & $70.5$                                                                                  & -                         & $67.7$                      \\ \hline
\multicolumn{7}{c}{\cellcolor[HTML]{EFEFEF}\textit{\textbf{Ours}}}                                                                                                                                                                                                                                                                                                                                                        \\ \hline
\rowcolor[HTML]{F8E4E2} 
GUI-Spotlight (Init. Qwen2.5-VL-7B)                     & $47.3$                                                                       & $50.0$                                                                             & $40.1$                                                                              & $20.2$                                                                                   & -                       & $35.6$                       \\
\rowcolor[HTML]{D5ECED}
GUI-Spotlight (Init. UI-TARS-1.5-7B)                      & $68.2$                                                                       & $60.6$                                                                             & $63.2$                                                                              & $45.6$                                                                                   & -                       & $62.7$                       \\ \hline
\end{tabular}}
\vspace{-10pt}
\end{table}

\subsection{General-Purpose GUI Visual Grounding}
OSWorld-G~\cite{xie2024osworld} comprises $564$ screenshots sourced from OSWorld~\cite{xie2024osworld}, covering a range of operating-system-level tasks such as file operations, application launching, text editing, and system configuration. It emphasizes general-purpose environments where agents must integrate recognition, layout reasoning, and manipulation in everyday workflows.

As shown in Table~\ref{tab:OSWorld-G}, GUI-Spotlight trained from UI-TARS-1.5-7B achieves an average accuracy of $62.7\%$, with particularly strong performance on text matching ($68.2\%$) and layout understanding ($63.2\%$). When initialized from Qwen2.5-VL-7B, the model raises the average score from $31.4\%$ to $35.6\%$, gaining substantially in element recognition (+$17.3$) and showing smaller improvements in text matching (+$1.7$) and manipulation (+$2.1$), with only a modest drop in layout understanding (–$1.8$). These results indicate that reinforcement learning with tool-augmented feedback provides clear benefits even when starting from a non-UI-specific backbone. Moreover, despite being trained on far fewer examples, the 7B-scale GUI-Spotlight remains competitive with 72B-scale models, supporting its robustness for diverse OS-level grounding tasks.

\subsection{GUI-Spotlight vs. training-free Iterative Inference}
\begin{wrapfigure}{r}{0.5\textwidth} % r=
  \centering
  \vspace{-13pt}
\includegraphics[width=0.98\linewidth]{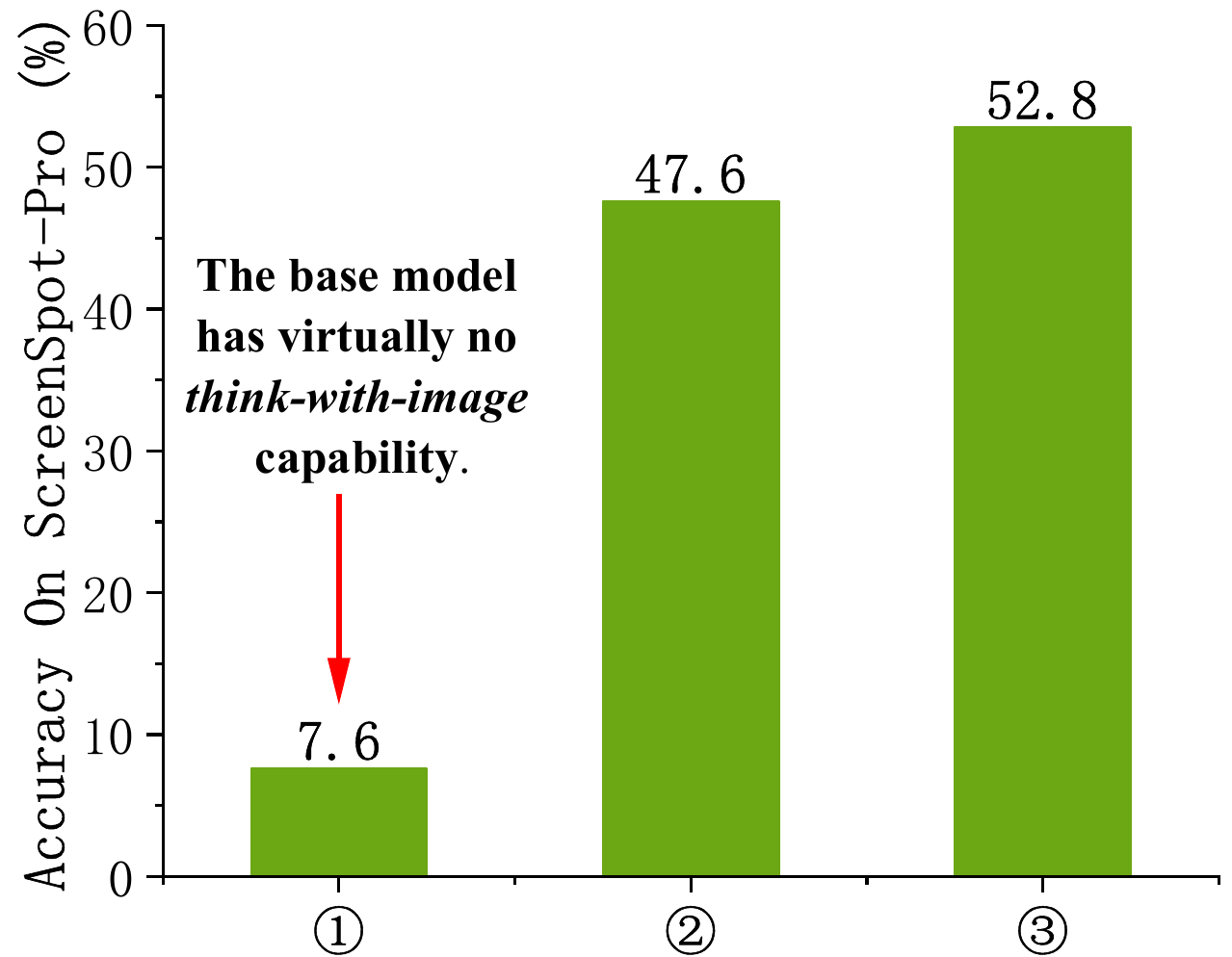}
  \caption{Comparison of multi-step reasoning strategies. UI-TARS-1.5-7B is used as the initial model: \ding{172} multi-turn conversational inference; \ding{173} repeated single-turn inference; \ding{174} GUI-Spotlight.}
  \label{fig:ablation}
  \vspace{-15pt}
\end{wrapfigure}
GUI-Spotlight performs multi-step reasoning at inference time. To quantify its gains over training-free iterative inference, we conduct an ablation study comparing GUI-Spotlight with two baselines:
\ding{172} Multi-turn conversational inference: we use the same multi-round tool prompts as GUI-Spotlight and, after each turn, append the executed tool outputs to the dialogue history.
\ding{173} Repeated single-turn inference: following the setting in InfantAgent-Next~\cite{lei2025infantagent} for vision models, after the first attempted click we crop a $700\times450$-pixel region centered at the predicted coordinates and continue issuing clicks within this region in subsequent attempts.

Our results show that the model initially has virtually no multi-step reasoning capacity. After training, however, the multi-step reasoning model attains higher accuracy than a baseline that iterates single-turn steps—at each step it performs one click, crops a local region centered on that click, feeds the cropped image back into the model, and then re-locates on the crop, repeating this procedure multiple times. This demonstrates a substantive post-training gain in GUI-Spotlight.

\section{Conclusion}
We introduced \textsc{GUI-Spotlight}, a \textit{think-with-image} visual grounding model that coordinates multiple tools through a stabilized GSPO-based reinforcement learning procedure. With only $18.5K$ training samples, it attains $52.8\%$ on \textsc{ScreenSpot-Pro} and $23.4\%$ on UI-Vision, remaining competitive with substantially larger models. Beyond raw accuracy, our multi-tool RL design improves training stability and sample efficiency, and our comprehensive documentation (including negative results) offers practical guidance for building agentic grounding models with coordinated tool use.

% We proposed \textsc{GUI-Spotlight}, a novel GUI visual grounding model that thinks with the image by dynamically narrowing its focus like a spotlight and refining predictions through tool-augmented reinforcement learning. Equipped with specialized tools (\textit{crop}, \textit{extract}, \textit{find color}) and trained with dense coordinate-level rewards and verifiable feedback, the model can iteratively localize fine-grained targets with high precision. Despite using only 18.5K curated samples, our 7B-scale model achieves 52.8\% accuracy on ScreenSpot-Pro, surpassing all existing 7B baselines and rivaling 32B and 72B models. These results establish tool-augmented reinforcement learning as a data-efficient and generalizable paradigm for developing practical GUI agents capable of reliable interaction in high-resolution, real-world environments.

\bibliography{iclr2026_conference}
\bibliographystyle{iclr2026_conference}

\appendix
\section{Appendix}
\label{appedix}
\subsection{Three Stages training hyperparameter}
\label{apd:3_stages_training}

The specific parameters used in the first stage of training are listed in Table~\ref{tab：stage_1_hype}. The parameters used in the second and third stage of training are listed in Table~\ref{tab：stage_2_3_hype}.

\begin{table}[ht]
\centering
\small
\caption{Training hyperparameters for Stage $1$}
\label{tab：stage_1_hype}
\begin{tabular}{cc}
\toprule
\textbf{Hyperparameter} & \textbf{Value} \\
\midrule
Finetuning type & \texttt{full} \\
Freeze vision tower & \texttt{true} \\
Freeze multimodal projector & \texttt{true} \\
Freeze language model & \texttt{false} \\
Precision & \texttt{bf16} \\
Learning rate & \texttt{1.0e-5} \\
LR scheduler & \texttt{cosine} \\
Warmup ratio & \texttt{0.1} \\
Per-device batch size & \texttt{12} \\
Gradient accumulation steps & \texttt{2} \\
Effective global batch & $12 \times 2 \times N_{\mathrm{GPU}}$ \\
Training epochs & \texttt{1.0} \\
Max sequence length & \texttt{10000} \\
Distributed/parallel & \texttt{DeepSpeed ZeRO-3} \\
\bottomrule
\end{tabular}
\label{tab:stage1-hparams}
\end{table}

\begin{table}[ht]
\centering
\small
\caption{Stage $2$ and $3$ training hyperparameters.}
\label{tab：stage_2_3_hype}
\begin{tabular}{cc}
\toprule
\textbf{Hyperparameter} & \textbf{Value} \\
\midrule
Learning rate & \texttt{1e-6} \\
LR scheduler & \texttt{constant\_with\_warmup} \\
Warmup steps & \texttt{10} \\
Training epochs & \texttt{1} \\
Temperature & \texttt{1.0} \\
Clip range $\epsilon$ & \texttt{0.2} \\
Clip range high $\epsilon_{\text{high}}$ & \texttt{0.28} \\
Precision & \texttt{bf16} \\
Max grad norm & \texttt{0.01} \\
Iterations per run & \texttt{2} \\
KL coefficient $\beta$ & \texttt{0.00} \\
Max prompt length & \texttt{1024} \\
Max completion length & \texttt{4096 (stage 2)/15000 (stage 3)} \\
Per-device train batch & \texttt{6} \\
Gradient accumulation steps & \texttt{16} \\
Num generations & \texttt{6} \\
\bottomrule
\end{tabular}
\label{tab:stage2-hparams}
\end{table}

\subsection{Training parameters for algorithmic exploration}
\label{apd:alg_exploration}

The specific training parameters for algorithmic exploration are listed in Table~\ref{tab：alg_exp}. By default, we use $\epsilon = 0.2$. In the \textit{clip-higher} setting, we set 
$\epsilon_{\text{low}} = 0.2$ and $\epsilon_{\text{high}} = 0.28$. 
The KL coefficient is $\beta_{\mathrm{KL}} = 0.01$ by default, 
and $\beta_{\mathrm{KL}} = 0$ in the \textit{no-KL} ablation.

\begin{table}[ht]
\centering
\small
\caption{Algorithmic exploration training hyperparameters.}
\label{tab：alg_exp}
\begin{tabular}{cc}
\toprule
\textbf{Hyperparameter} & \textbf{Value} \\
\midrule
Learning rate & \texttt{1e-6} \\
LR scheduler & \texttt{constant\_with\_warmup} \\
Warmup steps & \texttt{10} \\
Training epochs & \texttt{1} \\
Temperature & \texttt{1.0} \\
Precision & \texttt{bf16} \\
Max grad norm & \texttt{0.01} \\
Iterations per run & \texttt{2} \\
Max prompt length & \texttt{1024} \\
Max completion length & \texttt{4096 (stage 2)/15000 (stage 3)} \\
Per-device train batch & \texttt{6} \\
Gradient accumulation steps & \texttt{16} \\
Num generations & \texttt{6} \\
\bottomrule
\end{tabular}
\label{tab:stage2-hparams}
\end{table}

\subsection{Training and evaluation details for Experiments}
\label{apd:Experiments}

\noindent \textbf{Hardware} Experiments were conducted on a multi-GPU server with $8×$ NVIDIA H$200$ ($144$ GB HBM$3$e each), interconnected via NVLink (NV$18$ links across all pairs. The GPUs ran with NVIDIA driver $575.57.08$ and CUDA $12.9$ (MIG disabled, persistence mode enabled). The host is a dual-socket AMD EPYC 9454 machine ($2×48$ cores, $192$ threads total) with ~$1.5$ TiB system memory. We conducted all training and evaluation on GPUs $4–7$.

\noindent \textbf{Hyperparameters}
The training phase used the same hyperparameters as the three-stage training procedure (Appendix~\ref{apd:3_stages_training}). For evaluation, we employed the \texttt{vLLM} library; specifically, we set \texttt{tensor\_parallel\_size}=1, \texttt{gpu\_memory\_utilization}=0.95, \texttt{max\_model\_len}=30000, \texttt{max\_tokens}=19263, \texttt{temperature}=0, \texttt{top\_p}=1.0, and a batch size of 64.

\noindent \textbf{Prompts}
We consistently use the following prompt for both training and evaluation.
\begin{lstlisting}[style=promptstyle, caption={Tool\_Prompt}]
You should use three tools to help you analyze the image and find the target coordinate:
1. **crop**: This tool allows you to crop a specific area of the image by specifying the top-left and bottom-right coordinates of the rectangle you want to crop.
2. **extract**: This tool allows you to extract one quarter of the image based on the specified horizontal and vertical positions (left, center, right for x-axis; top, center, bottom for y-axis).
3. **find_color**: This tool allows you to find a specific color in the image by providing the RGB values of the target color.
Example Usage:
<crop>(Image_0, (10, 20), (110, 100))</crop> # Crop a rectangle from Image_0 from (10, 20) to (110, 100)
<extract>(Image_0, left, top)</extract> # Extract the top-left quarter of Image_0
<find_color>(Image_2, (255, 0, 0))</find_color> # Find the red color in Image_2
Before each tool call, please enclose your reasoning within <think>...</think> tags. 
In the end, you should return your final answer using the <answer>...</answer> tag.
In the <answer> tag, you should return the image and the coordinate of the target object in the format (Image_X, (x, y)), where Image_X is the image containing the target object and (x, y) is the coordinate of the target object.
Here is an example of how to find the final target coordinate:
User: [Image_0 is displayed above] Could you identify the location of the Close button in this interface?

Assistant: <think>I would like to zoom in on the menu icon at the top-left corner to get a clearer view.</think>  
<crop>(Image_0, (10, 20), (110, 100))</crop>

User: [Image_1 is displayed above]  # (cropped image returned)xw

Assistant: <think>In this cropped image, I cannot see the Close button; I will use the `extract` tool to find the possible area where the Close button is located.</think>
<extract>(Image_0, left, top)</extract>

User: [Image_2 is displayed above]  # (possible area returned)

Assistant: <think>In this cropped image, I can see the approximate position of the Close button, it sits near the center of the region, slightly toward the lower-right. It is at (45, 60).</think>  
<answer>(Image_2, (45, 60))</answer>

Now, let's work on the real task:
[Image_0 is displayed below]
\end{lstlisting}

\end{document}